\pdfoutput=1
\documentclass[final]{cvpr}
\usepackage{listings}
\usepackage{times}
\usepackage{epsfig}
\usepackage{graphicx}
\usepackage{amsmath}
\usepackage{amssymb}
\usepackage{subfigure} 
\usepackage{algorithm}
\usepackage{float}
\usepackage{algorithmicx}
\usepackage{algpseudocode}
\usepackage{array}
\usepackage{cite}
\usepackage{textcomp}
\usepackage{xcolor}
\usepackage{caption}
\usepackage{nopageno}
\def\cvprPaperID{****} 
\def\confYear{CVPR 2021}

\begin{document}

\title{Dynamic-OFA: Runtime DNN Architecture Switching for Performance Scaling on Heterogeneous Embedded Platforms}

\author{Wei Lou\thanks{Authors contributed equally.} \and Lei Xun\footnotemark[1] \and Amin Sabet \and  Jia Bi \and Jonathon Hare \and Geoff V. Merrett\\
University of Southampton, UK\\
{\tt\small \{wl4u19, lx2u16, ms4r18, J.Bi, jsh2, g.merrett\}@southampton.ac.uk}
\and
}

\maketitle

\begin{abstract}
Mobile and embedded platforms are increasingly required to efficiently execute computationally demanding DNNs across heterogeneous processing elements. At runtime, the available hardware resources to DNNs can vary considerably due to other concurrently running applications. The performance requirements of the applications could also change under different scenarios. To achieve the desired performance, dynamic DNNs have been proposed in which the number of channels/layers can be scaled in real time to meet different requirements under varying resource constraints. However, the training process of such dynamic DNNs can be costly, since platform-aware models of different deployment scenarios must be retrained to become dynamic. This paper proposes Dynamic-OFA, a novel dynamic DNN approach for state-of-the-art platform-aware NAS models (\ie Once-for-all network (OFA)). Dynamic-OFA pre-samples a family of sub-networks from a static OFA backbone model, and contains a runtime manager to choose different sub-networks under different runtime environments. As such, Dynamic-OFA does not need the traditional dynamic DNN training pipeline. Compared to the state-of-the-art, our experimental results using ImageNet on a Jetson Xavier NX show that the approach is up to 3.5x (CPU), 2.4x (GPU) faster for similar ImageNet Top-1 accuracy, or 3.8\% (CPU), 5.1\% (GPU) higher accuracy at similar latency.
\end{abstract}

\section{Introduction}

\begin{figure}[!htb]

\subfigure[ ] 
{
	\begin{minipage}{3.7cm}
	\centering        
	\includegraphics[width=\linewidth]{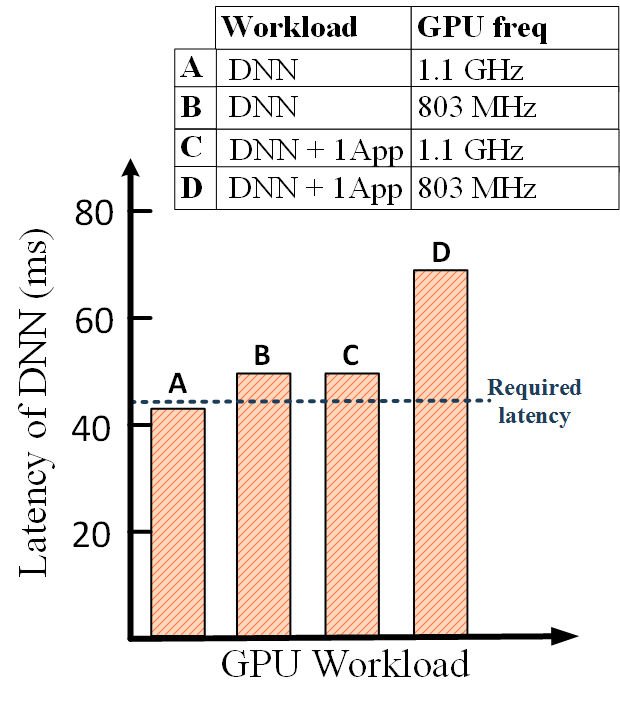}   
	\end{minipage}
}
\subfigure[  ] 
{
	\begin{minipage}{3.7cm}
	\centering     
	\includegraphics[width=\linewidth]{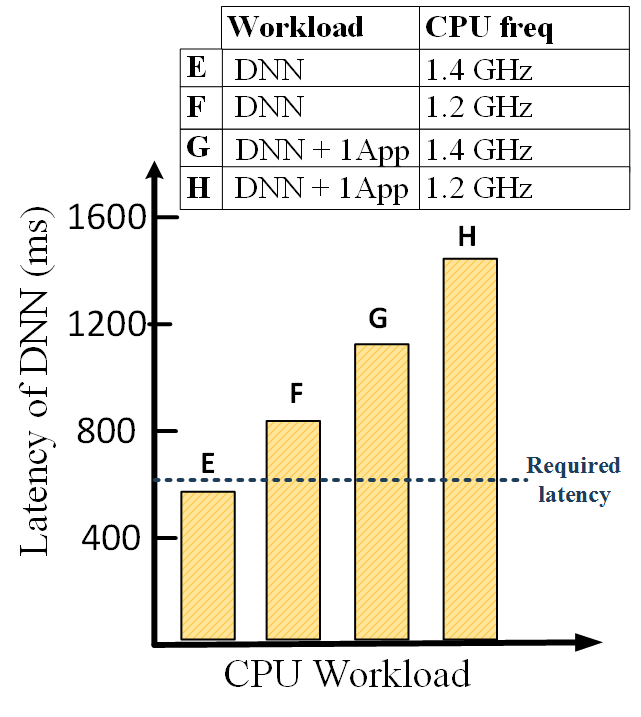}   
	\end{minipage}
}
 
\caption{Experimental results illustrating how inference latency constraints characterised at design-time can be violated at runtime by changes in available hardware resources and executing tasks. (a) denotes the latency on GPU and (b) on CPU. The tables above indicate the combinations of workload and frequency.} 
\label{fig1}
\end{figure}

\begin{figure*}[ht]
\includegraphics[width=1.0\linewidth]{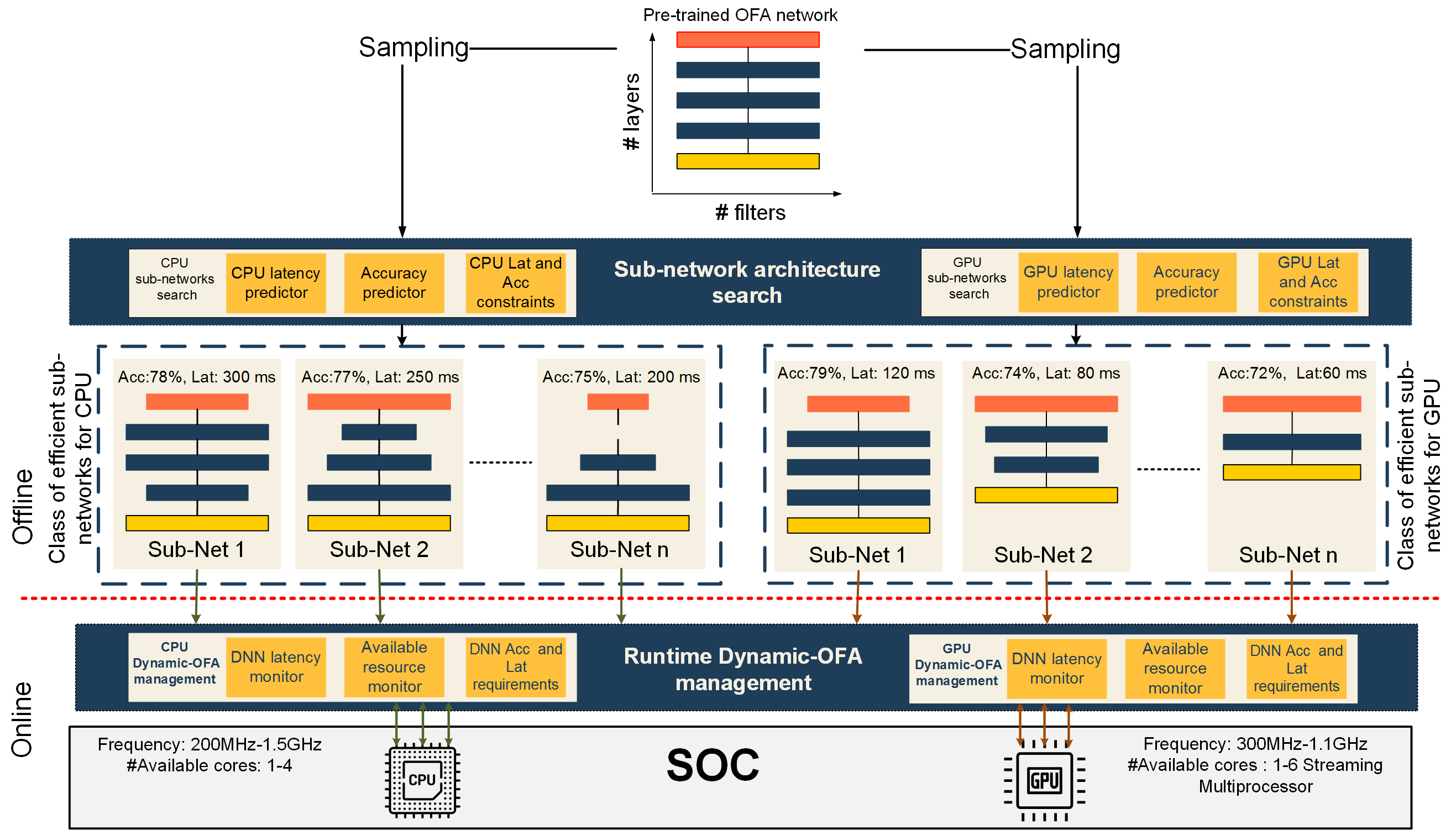}
\centering
\caption{The workflow of Dynamic-OFA. Using pre-trained OFA networks that contain $2 \times {10}^{19}$ sub-network architectures as the backbone, sub-network architectures are sampled from OFA for both CPU and GPU at the offline stage. These architectures have different performance (\eg latency, accuracy) and are stored in a look-up table to build a dynamic version of OFA without any additional training required. Then, at runtime, Dynamic-OFA selects and switches to optimal sub-network architectures to fit time-varying available hardware resources.}
\label{fig2}
\end{figure*}

Modern heterogeneous embedded systems typically execute multiple applications concurrently. DNNs are increasingly being executed on embedded platforms due to the superior performance in many applications such as computer vision and natural language processing. Compared to cloud-based solutions, DNN inference on embedded platforms brings many advantages in latency, privacy and connectivity. However, DNN inference is both computationally and memory intensive, making it a challenge to deploy on embedded platforms.

Many approaches have been proposed to reduce the computational demands of DNN inference, such as platform-aware filter pruning \cite{yang2018netadapt, he2018amc} and Neural Architecture Search (NAS) \cite{cai2019once, cai2018proxylessnas}. These approaches produce a static DNN architecture with fixed parameters for the target application performance requirements based on the measurement on a fixed hardware resources. However, since available hardware resources dynamically change at runtime, performance requirements can be violated \cite{xun2020optimising, reform}. Fig~\ref{fig1} illustrates these problems by using experimental results from a Jetson Xavier NX, where bar $A$ represents an optimized DNN model executing on all GPU cores at 1.1GHz to deliver a 50ms target latency. However, under the same target latency, the optimization at design-time is invalid if the operating frequency changes or the DNN shares GPU cores with other applications at runtime, as shown by bars $B$, $C$ and $D$. The same applies in the CPU case, shown by bars $E$, $F$, $G$ and $H$.

Since both software performance requirements and hardware resource availability can change dynamically at runtime \cite{xun2020optimising, reform}, various dynamic DNNs \cite{yu2018slimmable, yu2019universally, yu2019autoslim, yang2020mutualnet} have been proposed to address this issue. These dynamic DNNs contain various sub-networks that each have a different accuracy and latency. Given different available hardware resources and application requirements, different sub-networks can be selected. However, two problems still remain in the previous approaches. First, previous dynamic DNNs involve an additional training pipeline which retrains the backbone network (\eg MobileNet\cite{howard2017mobilenets, sandler2018mobilenetv2}, ResNet\cite{resnet}) from scratch. The total training would be costly in real deployment scenarios, since platform-aware pruning/NAS need to be applied before dynamic DNN process, and all model variants of different deployment scenarios have to be retrained to become dynamic. Second, prior works re-scale the network architecture to obtain sub-networks by either channel scaling \cite{yu2018slimmable, yu2019universally, yu2019autoslim, yang2020mutualnet} or layer scaling \cite{huang2017multi}. However, the most efficient DNN architectures on CPUs typically have more layers but fewer channels, while the opposite is true for GPUs \cite{cai2018proxylessnas}. Therefore, previous scaling methods limit the application of dynamic DNN on heterogeneous System on Chip (SoC) platforms, since modern SoCs integrate multiple computing elements including CPU, GPU and NPU. As a result, different dynamic DNNs are needed for different heterogeneous computing elements.

This paper proposes Dynamic-OFA, a novel approach to provide dynamic DNN architectures for different software and hardware resource requirements. Dynamic-OFA works with state-of-the-art platform-aware NAS methods (\ie Once-for-all network (OFA)) without requiring any additional model retraining. Moreover, to improve previous dynamic approaches, Dynamic-OFA scales the entire DNN architecture, including width, depth, filter sizes, and input resolutions to provide more efficient DNNs architectures for both CPU and GPU with one shared backbone.

Fig~\ref{fig2} shows the workflow of Dynamic-OFA with two main steps. In the first \textit{offline} step, different sub-networks are extracted from the OFA \cite{cai2019once} super-network. Accuracy and latency of extracted sub-networks are evaluated to find a family of efficient sub-networks on the Pareto-front for both CPU and GPU. In the second step, and during \textit{runtime}, Dynamic-OFA uses a runtime manager to switch between the optimal sub-network based on the runtime accuracy and latency requirements of the application, and available resources on the platform.

The contributions of this paper are:

\begin{enumerate}
\item A novel dynamic DNN approach that combines OFA with dynamic DNN. Compared to the state-of-the-art, our experimental results on Nvidia Jetson Xavier NX show that our approach is up to 3.5x (CPU), 2.4x (GPU) faster for similar ImageNet Top-1 accuracy, or 3.8\% (CPU), 5.1\% (GPU) higher accuracy at similar latency.
\item An improved search algorithm to efficiently search a family of sub-networks from the OFA super-network, by jointly considering inference accuracy and latency.
\item A runtime approach to dynamically switch between sub-networks to meet  time-varying performance constraints and/or available hardware resources.
\end{enumerate}

\section{Related work}
\paragraph{Neural Architecture Search (NAS)}
Many handcrafted efficient DNN models like MobileNet \cite{howard2017mobilenets, sandler2018mobilenetv2} and ShuffleNet \cite{ma2018shufflenet} achieve state-of-the-art performance. However, these models need to be further compressed to fit constraints of the target device using platform-aware compression techniques\cite{yang2018netadapt, he2018amc}. Since compression needs to be conducted for each new constraint, the time required for large-scale deployment is expensive. Furthermore, designing handcrafted DNN models needs expert knowledge and can be a challenging and time-consuming task. NAS automates architecture design, directly searching for the most efficient DNN architectures for target constraints. However, given new platform constraints, DNN models need to be researched and retrained; therefore, the required time is still prohibitive for large-scale deployments. For example, Tan \etal \cite{tan2019mnasnet} introduced MnasNet, a multi-objective NAS approach that uses reinforcement learning to find the architecture for maximizing accuracy and latency on target hardware platforms. MnasNet cost 40,000 GPU hours (Nvidia V100 GPU) to find a single DNN architecture, and this raises a significant issue since the number of combinations of hardware platforms and software performance targets are significant. Cai \etal \cite{cai2018proxylessnas} proposed ProxylessNAS, which uses weight-sharing and differentiable architecture search; however, the search cost for each model is still 200 GPU hours which is also significant for large scale deployments. 

Cai \etal \cite{cai2019once} proposed the Once-for-all network (OFA), a NAS approach that supports large-scale deployments. Only one DNN model (a super-network) needs to be trained, and enables $2 \times {10}^{19}$ sub-networks to be used for different software performance requirements on different hardware platforms. While OFA offers a number of significant advantages, the available hardware resources in modern SoCs, containing heterogeneous computing elements that use energy-efficient features such as DVFS and task mapping, vary dynamically. OFA is not a dynamic DNN by default, since its search process is still too time-consuming (\eg hours on Nvidia Jetson Xavier NX). This prevents \textit{real-time} architecture switching at runtime, and motivates us to propose a general approach for building dynamic DNNs for OFA super-networks.

\paragraph{Dynamic DNNs}
Dynamic DNNs can switch the DNN architecture to fit time-varying available hardware resources and software performance requirements. A variety of approaches have been proposed to change the width of the backbone networks, like ‘Slimmable’ \cite{yu2018slimmable}, ‘Universally Slimmable’ \cite{yu2019universally} and ‘AutoSlim’ \cite{yu2019autoslim}. These models can run different active channels and achieve instant and adaptive accuracy-latency trade-offs. MutualNet \cite{yang2020mutualnet} shows improved performance by adding input resolutions with width as switchable dimensions. For different constraints, different sub-networks with varying widths and resolutions can be chosen and built as a query table. Approaches such as MSDNet \cite{huang2017multi} can change the depths of the backbone networks. However, the most efficient models for different hardware resources are usually different in architecture. For example GPUs prefer shallow and wide DNN architectures with early pooling, while CPUs prefer deep and narrow DNN architectures with late pooling \cite{cai2018proxylessnas}. The switchable dimensions for previous approaches are either width or depth; therefore, they cannot provide the most efficient architecture for both CPU and GPU on the modern SoCs. The Dynamic-OFA approach presented in this paper switches in four dimensions (width, depth, filter sizes, and input resolutions) to provide efficient architectures for both GPUs and CPUs by using a share backbone model.

\section{Dynamic-OFA}
As illustrated in Fig \ref{fig2}, Dynamic-OFA use the following process: a pre-trained OFA model \cite{cai2019once} is used as the backbone (super-network), as introduced in Section \ref{sub3_a}. As an offline process, sub-networks with different latency and accuracy are sampled from the pre-trained super-network at design-time (Section \ref{sub3_b}). The sampling process is conducted separately for CPU and GPU because the most efficient sub-network architectures are different for heterogeneous computing resources. The batch-normalization (batch-norm) parameters of each sub-network are recalculated and stored (Section \ref{sub3_c}). At runtime, the sampled sub-network architecture can be switched to meet latency and accuracy requirements on time-varying available hardware resources (Section \ref{sub_d}). No additional training is needed, and since all sub-networks share the same OFA model as the backbone, only a single model needs to be stored on the device.

\subsection{Backbone Network: Once-For-All (OFA) } \label{sub3_a}
The OFA network \cite{cai2019once} supports $2 \times {10}^{19}$ sub-networks with different sizes by a single super-network, covering four dimensions of the DNN architecture: depth, width, kernel, and input resolutions. OFA uses progressive shrinking to train the super-network model. Progressive shrinking optimizes super-network parameters such that each supported sub-network maintains almost the same level of accuracy as independently training a network with the same architecture configuration. After training the super-network model, OFA uses an evolutionary search algorithm \cite{real2019regularized} to find sub-network architectures in a super-network model with different accuracy and latency trade-offs. OFA only requires a single set of parameters to store all sub-networks, since all sub-networks share the same parameters. OFA is able to select sub-network architectures for different hardware platforms, and then calibrates the batch-norm parameters for those selected sub-networks. The process of searching for sub-networks and calibrating batch-norm parameters take minutes on a GPU or hours on a CPU\footnote{The platform is Nvidia Jetson Xavier NX. The time includes measure the accuracy of selected sub-network on ImageNet 50k validation set, since the accuracy prediction is not accurate enough, more details in section \ref{sub3_b}. Batch-norm is calculated on 2000 images.}. The actual search time at runtime depends on how tight the constraints are, and how the search problem shares hardware resources with other applications, and hence OFA cannot be used to adapt to resource changes at runtime. 

Our approach identifies a family of efficient sub-networks on the Pareto-front for each computation element in a heterogeneous platform, and pre-calculates batch-norm parameters for those sub-networks offline. This allows switching the network architecture at runtime. Furthermore, since Dynamic-OFA conducts all the sub-network search at design-time on a server, the search cost is dramatically reduced to that of the single search cost, multiplied by the number of sub-networks. 

\begin{algorithm} [ht]
	\renewcommand{\algorithmicrequire}{\textbf{Input:}}
	\renewcommand{\algorithmicensure}{\textbf{Output:}}
	\small
    \caption{Random Search Algorithm of Dynamic-OFA.}
    \begin{algorithmic}[1]
    \Require {maximum accuracy constraint ($Acc_{max}$), accuracy suitable range ($Acc_{r}$), latency ($Lat$), maximum latency constraint ($Lat_{max}$), initial latency ($Lat_{init}$), extracted sub-networks ($Parents$), latency constraint increment ($Lat_{add}$), $n$ iterations of searching ($itr_{n}$)}, maximum iterations number ($itr_{max}$)
    \Ensure {qualified network architecture ($Arch$), predicted accuracy ($Acc_{p}$), predicted latency ($Lat_{p}$)}
    \Function {Random\_Search}{}
        \State $Lat \gets Lat_{init}$
        \State $i= 0$ 
        \While {$i<itr_{max}$}{
        \If {$i\% itr_{n}=0$ \textbf{and} $Lat<Lat_{max}$}
            \State {$Lat \gets Lat+Lat_{add}$}
        \EndIf
            \State $Arch = Random\_Arch()$
            \State $Acc_{p} = Predict\_Accuracy(Arch)$
            \State $Lat_{p} = Predict\_Latency(Arch)$
        \If {$Acc_{p} > Acc_{max}-Acc_{r}$ \textbf{and} $Lat_{p}<Lat$}
            \If{$Arch$ \textbf{not in} $Parents$}             
                \State \Return{$Arch, Acc_{p}, Lat_{p}$}
            \EndIf
        \EndIf
            \State $i \gets i+1$
        \EndWhile }
    \EndFunction
    \end{algorithmic}
\label{algo1}
\end{algorithm}

\subsection{Optimal Sub-network Architecture Search} \label{sub3_b}

OFA's search is solely based on the latency constraint, as it selects sub-networks with the highest accuracy that meets a latency constraint. It uses a combination of a random search and evolutionary search: (1) a random search algorithm finds sub-networks that meet the latency constraints; (2) the evolutionary search performs refinement to mutate those random selected sub-network configurations, and then keeps the one which has the highest accuracy. Although this approach works for a single sub-network architecture search, it is not efficient for searching a family of sub-networks. To build dynamic-DNNs, multiple sub-network architectures with different accuracy-latency trade-offs are required to be dynamically switched at runtime. If we only use latency as the constraint (which is the original OFA search method), the random search algorithm often outputs the sub-networks with the same accuracy for different latency constraints. For example, a 75\% accuracy, 30 ms model is searched under 30 ms, 40 ms and 50 ms constraints, but ideally we want models with higher accuracy when the latency is relaxed. The OFA algorithm only requires the latency of searched sub-networks to be lower than constraint but not for accuracy, and this prevents us from getting the Pareto trade-off curve. In our random search algorithm \ref{algo1}, by adding accuracy as a hard constraint, it forces the search algorithm to find sub-networks under certain latency constraints and have better accuracy (\eg 75\% accuracy under 30 ms, 76 \% accuracy under 40 ms, 77\% accuracy under 50 ms etc).

In Algorithm \ref{algo1}, accuracy and latency predictors are used to reduce the time cost of the search process. We use the original accuracy predictor that comes with the pre-trained OFA model; it is a three-layer neural network trained using 5000 network architectures and their accuracy data. However, since we want the dynamic DNN to have 1\% accuracy steps between sub-networks, the accuracy of all searched models are re-measured on the ImageNet 50K validation set since the original OFA accuracy predictor is not accurate enough (normally 2-4\% higher than our measurement). We adapted the latency predictor in OFA for Nvidia Jetson Xavier GPU and CPU as a heterogeneous platform. A lookup table is used to record the latency for different DNN computing operations such as convolution, pooling, input and output. The latency of all operations is profiled at design-time by timing the execution time of all operations at all size variants. Moreover, the separate lookup table is built for different hardware platforms and different computing cores on these platforms (\eg CPU, GPU). The latency of a sub-network can be predicted by summing up the latency of sub-network operations. The latency are also re-measured on devices for more accurate data.

To search for sub-networks, the latency constraint is initialized with a user-defined value $Lat_{init}$, shown in algorithm \ref{algo1} line 2. An accuracy constraint is also fixed at a user-defined range $[Acc_{max}-Acc_r, Acc_{max}]$. Then,  \textit{RANDOM\_SEARCH} starts searching for sub-networks by randomly generating sub-network architectures $Random\_Arch()$ (line 8), \ie different depths, width, kernel sizes and input resolutions. $Predict\_Latency()$ and $Predict\_Accuracy()$ predict the latency and accuracy of each sub-network in lines 9 and 10. If none of the found architectures provide a latency of less than $Lat$ and an accuracy in the range of $[Acc_{max}-Acc_r, Acc_{max}]$ for n iterations of searching ($itr_{n}$), the latency constraint is relaxed by increasing $Lat_{add}$ (line 6). However, this relaxation cannot exceed a maximum threshold $Lat_{max}$. The discovered network architecture is stored in $Parents$ (line 13), which is subsequently used for mutation operations of the evolutionary search algorithm to search for the best sub-network.

\begin{figure}[t]
\begin{center}
   \includegraphics[width=0.95\linewidth]{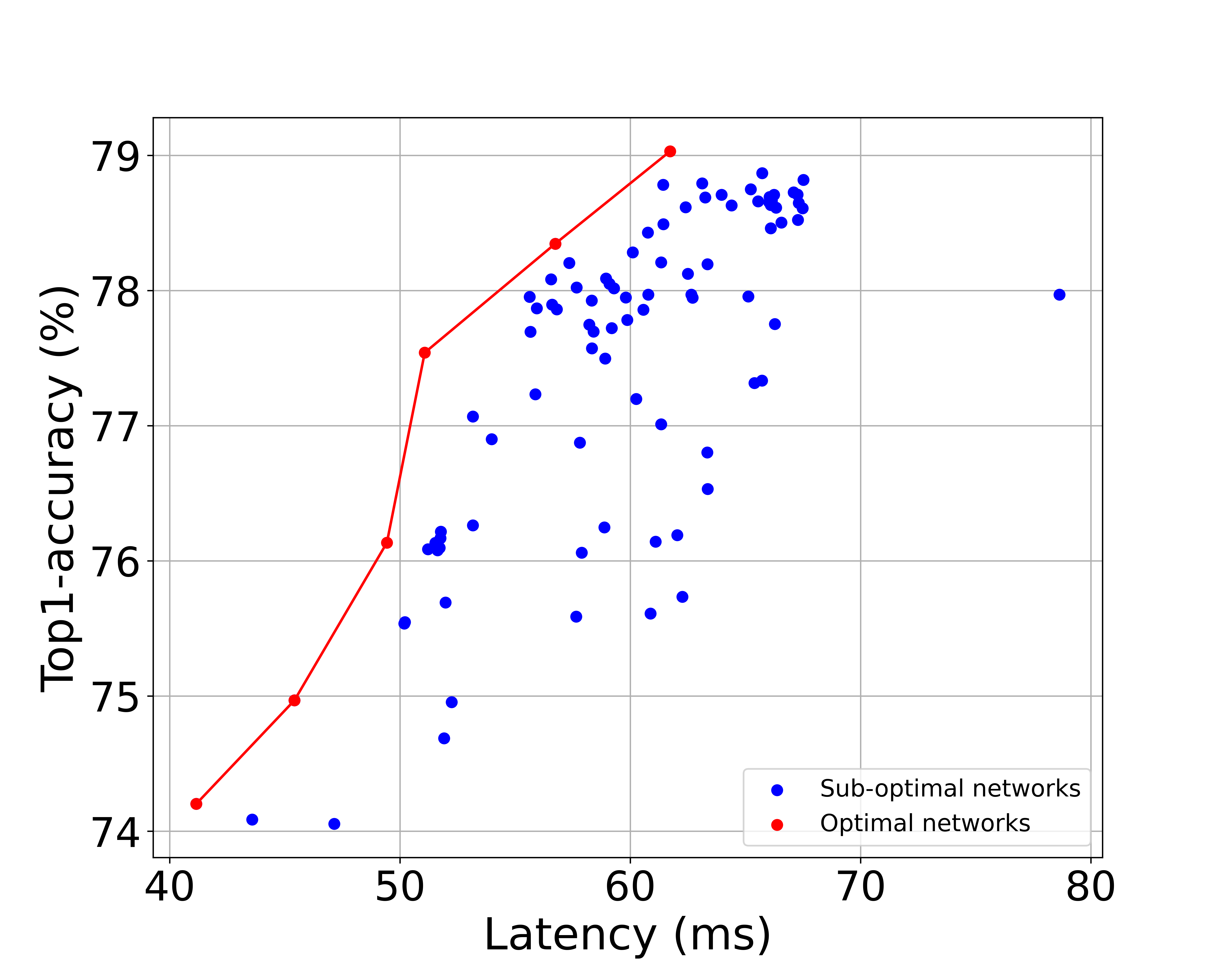}
\end{center}
   \caption{The accuracy and latency of potential sub-network architectures. The optimal architectures are on the Pareto curve.}
\label{fig3}
\end{figure}

We use the state-of-the-art pre-trained OFA \cite{cai2019once} super-network model as the backbone network for image classification. This model uses MobileNet v3 \cite{howard2019searching} as the fundamental network architecture and is trained with the progressive shrinking approach \cite{cai2019once} to enable $2 \times {10}^{19}$ sub-networks. A random search, Algorithm \ref{algo1}, is used within the evolutionary search algorithm\footnote{The same evolutionary search algorithm that the OFA used.} \cite{real2019regularized} to search for potential efficient sub-networks. It is notable that for given accuracy and latency constraints, numerous sub-network architectures can be found in the search process for each computation element, \eg CPU, GPU in a heterogeneous platform. However, most architectures are sub-optimal because they provide lower accuracy or higher latency. The optimal sub-network architectures are on the Pareto curve of accuracy-latency scatter plot, and they are selected for building the Dynamic-OFA. Fig \ref{fig3} shows the accuracy-latency scatter plot of extracted sub-networks for the GPU of Nvidia Jetson Xavier NX platform. Among obtained sub-networks, only those located on the Pareto-front (red line) are considered as a family of optimal sub-networks for use by Dynamic-OFA.

\begin{figure*}[!htb]
\subfigure[ GPU ] 
{
	\begin{minipage}{8.9cm}
	\centering        
	\includegraphics[width=\linewidth]{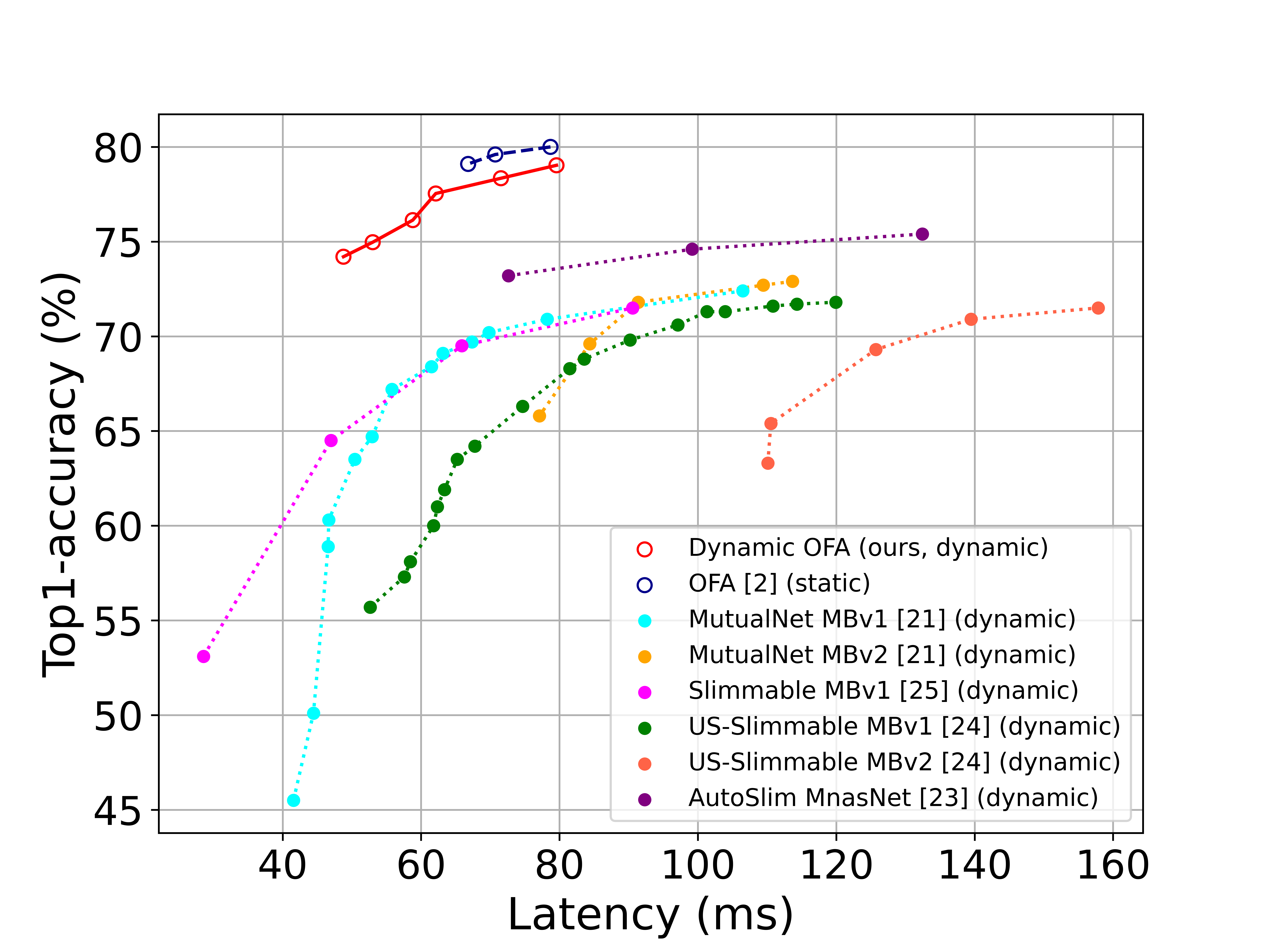}   
	\end{minipage}
}
\subfigure[ Single-core CPU ] 
{
	\begin{minipage}{8.9cm}
	\centering     
	\includegraphics[width=\linewidth]{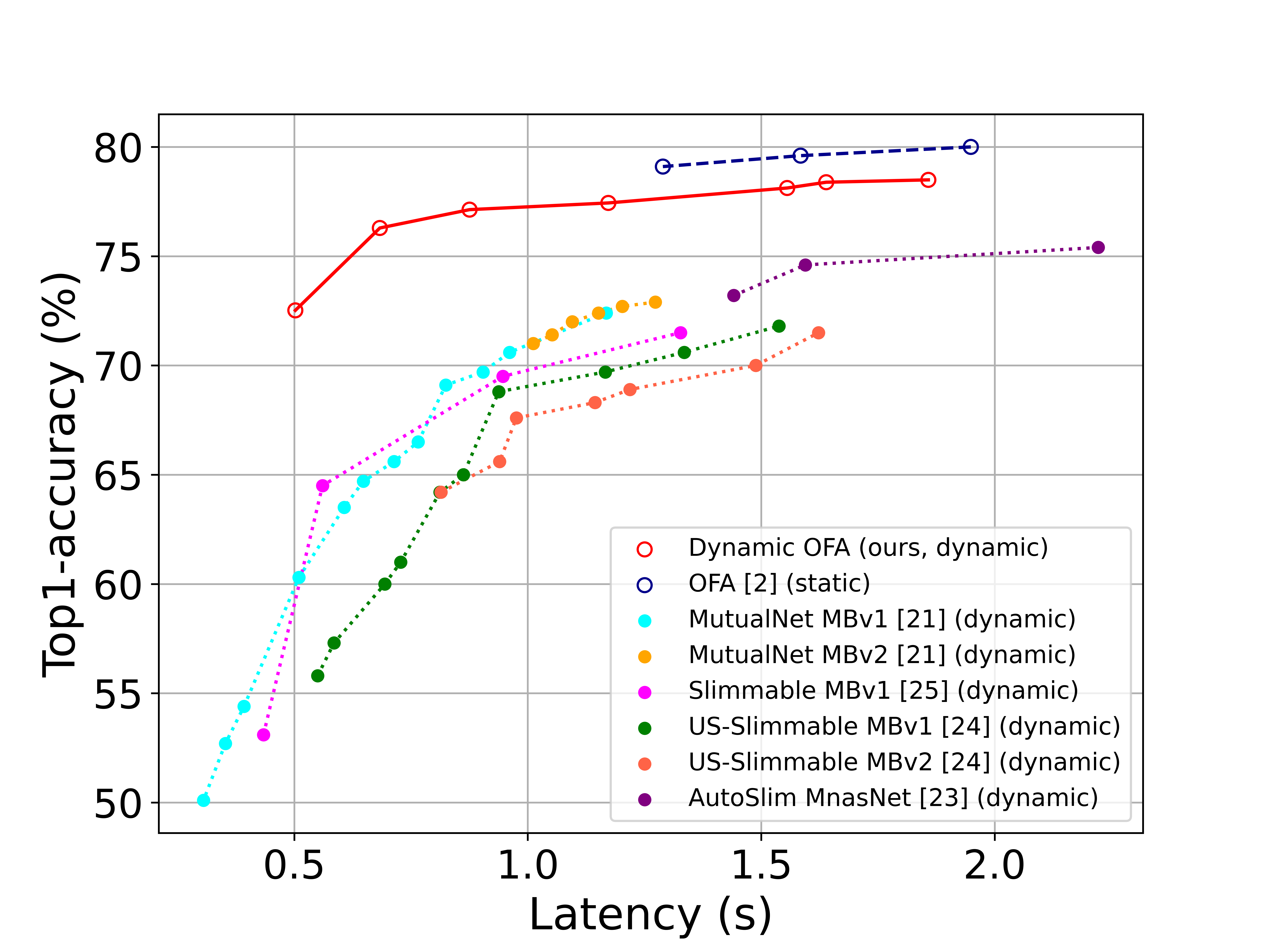}   
	\end{minipage}
}
 
\caption{Experimental results of Dynamic-OFA's accuracy-latency trade-offs on the a) GPU and b) CPU of the Nvidia Jetson Xavier NX. State of the art approaches (shown in different colours) are also plotted, including static \cite{cai2019once} and dynamic DNNs \cite{yu2018slimmable,yu2019universally, yu2019autoslim, yang2020mutualnet}. Dynamic-OFA is 2.4x (GPU) and 3.5x (CPU) faster (at similar accuracy) or has 5.1\% (GPU) and 3.8\% (CPU) higher Top-1 ImageNet accuracy (at similar latency) than AutoSlim-MnasNet \cite{yu2019autoslim}.} 
\label{fig4}
\end{figure*}

\begin{figure}[t]
\begin{center}
   \includegraphics[width=1.05\linewidth]{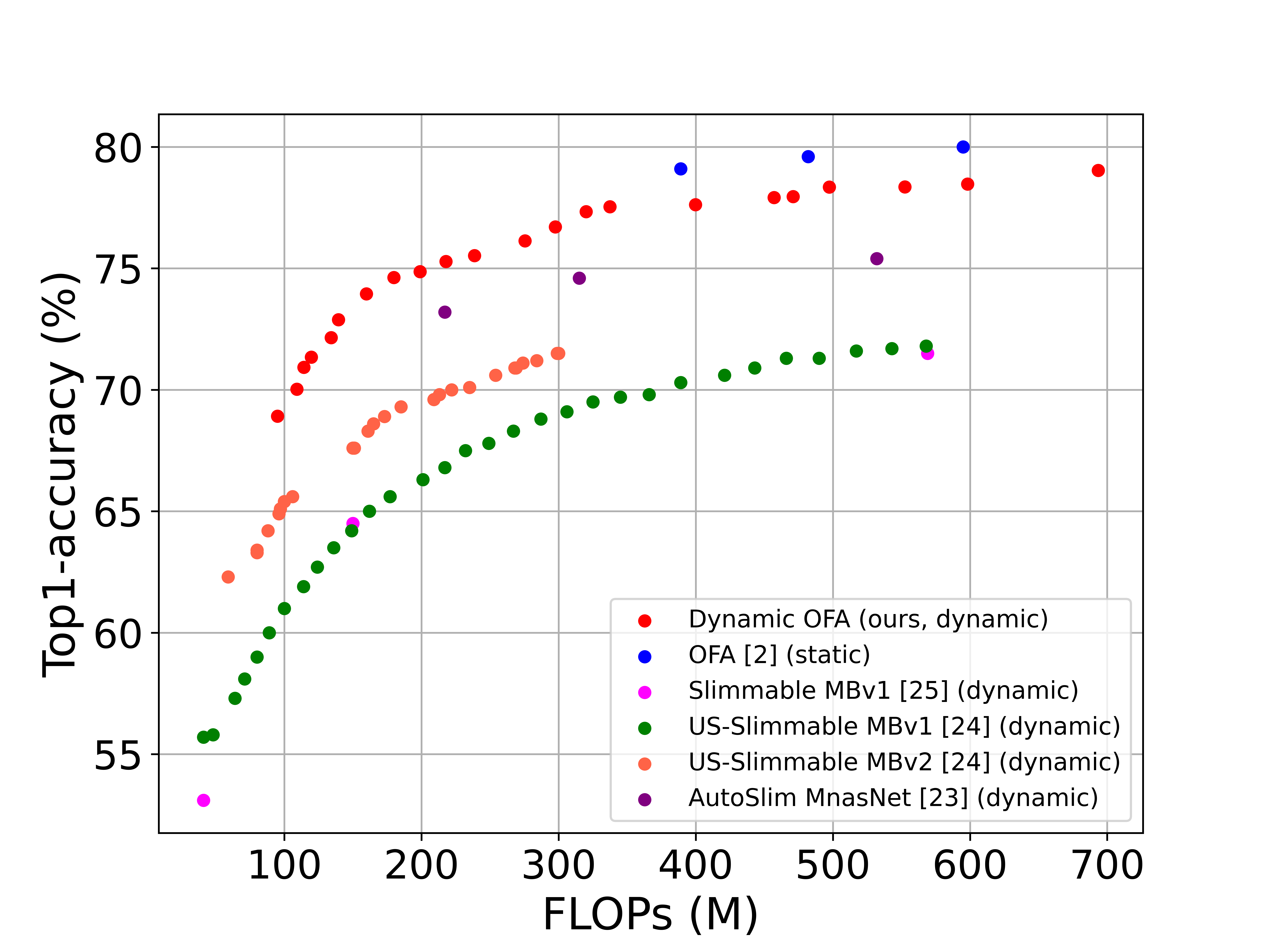}
\end{center}
   \caption{Experimental results of Dynamic-OFA's accuracy-FLOPs trade-offs. Compare with state-of-the-art approaches, Dynamic-OFA achieves up to 50\% FLOPs reduction (at similar accuracy) and 2.95\% higher Top-1 ImageNet accuracy (at similar FLOPs) than AutoSlim-MnasNet \cite{yu2019autoslim}.}
\label{fig5}
\end{figure}
\subsection{Batch-norm Calibration} \label{sub3_c}
Searching sub-networks at runtime is computationally expensive. Therefore, to make OFA dynamic at runtime, the search algorithm is run offline, and the optimal sub-network configurations are stored as a lookup table to achieve fast architecture switching at runtime. Moreover, the batch-norm parameters \cite{Sergey2015BN} differ for each sub-network architecture and need to be recalculated for different sub-networks. However, the calibration can take minutes on GPU and hours on CPU, which is unacceptable for real-time DNN architecture switching at runtime. Since we have selected a small number of optimal sub-network architectures, the batch-norm parameters can be pre-calculated at design-time. At runtime, sub-network architectures can be switched to meet different performance requirements on time-varying available hardware resources. The pre-stored batch-norm parameters (about 2KB for each sub-network) can be loaded, significantly reducing the architecture switching time. 

\subsection{Runtime Architecture Switching} \label{sub_d}
At runtime, sub-network architectures of Dynamic-OFA can be switched to meet different performance requirements on time-varying available hardware resources. Because of the stored sub-network architectures and their batch-norm parameters, Dynamic-OFA supports real-time architecture switching for the dynamic computing environments. We consider two operating scenarios:

\begin{enumerate}
\item When a single dynamic-OFA runs on the device, a look-up table is used to directly find the sub-network `level' to meet different user-defined accuracy and latency constraints. This is a similar approach to previous dynamic DNNs like Slimmable \cite{yu2018slimmable,yu2019universally,yu2019autoslim} and MutualNet \cite{yang2020mutualnet}.

\item When two workloads share the same GPU resources (\eg one dynamic-OFA with another workload, or two dynamic-OFAs), a reactive control approach is used instead of a look-up table, due to the increased state space. The latency of the current Dynamic-OFA model is measured over a certain time interval using a sliding window. The RTM continually monitors the latency of all Dynamic-OFA workloads at runtime. When a latency constraint violation occurs, the RTM gradually changes sub-network levels while observing whether latency constraints are subsequently met. Although the design-time profiled look-up table is not used, the trade-off between sub-networks still holds.
\end{enumerate}

The overhead of the RTM includes the monitoring of latency and the calculation of average latency, neither of which are computationally expensive in comparison to the task being monitored. Detailed overhead measurement will be shown in the next section.

\section{Experimental evaluation}

Dynamic-OFA is developed for both the CPU and GPU of the Nvidia Jetson Xavier NX platform, and the accuracy and latency are empirically measured. The results are compared with both state-of-the-art static OFA backbone \cite{cai2019once} and dynamic DNNs \cite{yu2018slimmable, yu2019universally, yu2019autoslim, yang2020mutualnet}. We have also executed other applications (\eg online model training\cite{PlantCLEF} or a 2nd Dynamic-OFA) alongside our Dynamic-OFA, to demonstrate runtime management of the sub-network architectures for accommodating both applications on the same GPU.

\subsection{Experimental Setup}
We deploy and evaluate Dynamic-OFA on the Nvidia Jetson Xavier NX. The platform has a 384-core GPU, a 6-core CPU, and 8 GB of unified memory. The frequency is locked at the maximum\footnote{Due to the intensive nature of DNN workloads, current DVFS governors will typically operate the SoC at the maximum frequency.}. A single-core CPU is used for all CPU experiments since the program is single-threaded. We use the open-source pre-trained OFA model (MobileNet v3, $w=1.2$ \cite{cai2019once}) as the backbone for Dynamic-OFA. We evaluate the accuracy of Dynamic-OFA on the ImageNet\cite{imagenet} 50K validation images for a more accurate measurement than OFA's accuracy predictor.

\begin{figure}[t]
\begin{center}
   \includegraphics[width=0.9\linewidth]{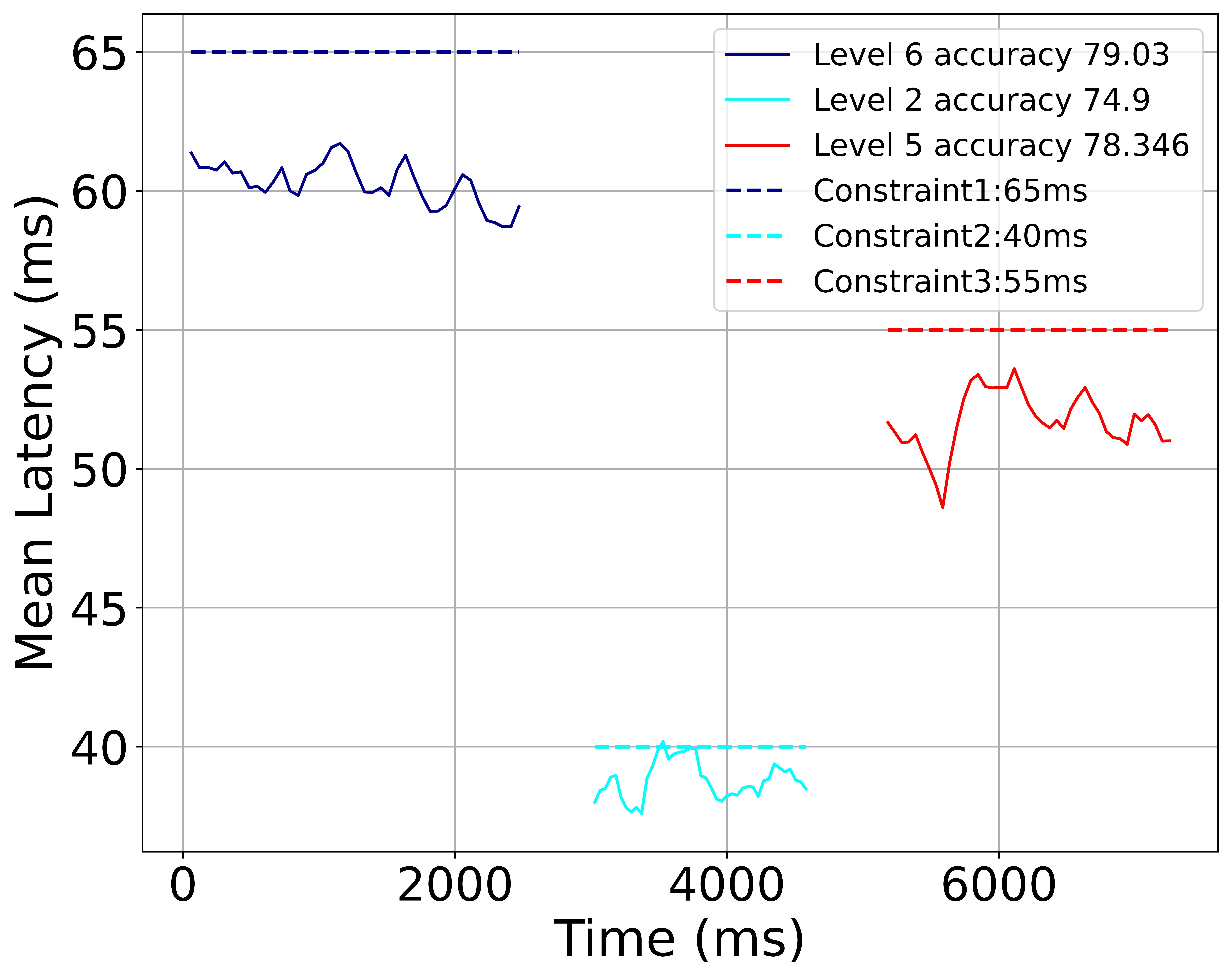}
\end{center}
   \caption{Runtime management of dynamic-OFA to meet the dynamic latency constraints. Latency constraints are denoted by dashed lines. Different sub-networks with different accuracies are denoted using different colours.}
\label{fig6}
\end{figure}

\subsection{Top-1 Accuracy and Latency}
Dynamic-OFA contains six optimal sub-networks for the GPU, which span a 20-100 ms latency range (Fig \ref{fig4}a). The same model also contains seven optimal sub-networks for the CPU, spanning a range from 500-3000 ms (Fig \ref{fig4}b). The search range for Top-1 accuracy is 70\% to 85\% for both CPU and GPU. In this paper, we denote sub-networks using ``levels.'' Different levels have different latency and accuracy. For example, level 1 is the sub-network architecture that has the lowest latency and accuracy. On GPU, level 6 is the architecture that has the highest latency and accuracy, whereas, on CPU, level 7 has the highest latency and accuracy. 

Dynamic-OFA achieves up to 79\% Top-1 accuracy on ImageNet; close to the 80\% accuracy of the static OFA backbone \cite{cai2019once} which is considered the state-of-the-art under mobile settings ($<$600M MACs). The reason our Dynamic-OFA has 1\% lower accuracy is because OFA applies extra fine-tuning steps for its individual model, whereas our model cannot be easily fine-tuned since it contains 13 different sub-networks. Compared to state-of-the-art dynamic DNNs \cite{yu2019autoslim}, Dynamic-OFA is up to 2.4x (GPU) and 3.5x (CPU) faster (at a similar accuracy) or has 5.1\% (GPU) and 3.8\% (CPU) higher Top-1 ImageNet accuracy (at a similar latency). Furthermore, since previous dynamic DNNs are designed using FLOPs rather than latency, they can access less hardware information during design-time. Hence we also compared our Dynamic-OFA with them on a accuracy-FLOPs trade-off curve (Fig \ref{fig5}). Dynamic-OFA achieves up to 50\% FLOPs reduction (at similar accuracy) and 2.95\% higher Top-1 ImageNet accuracy (at similar FLOPs) than prior art.

\begin{table}
\centering  
\setlength{\tabcolsep}{9mm}
\resizebox{\columnwidth}{!}{
\begin{tabular}{c|c}  
\hline  
\textbf{Model} &  \textbf{Time} \\  
\hline  
Static OFA\cite{cai2019once} & minutes to hours\\  
\hline  
MutualNet-MBv2 \cite{yang2020mutualnet}& 17 ms\\
\hline  
AutoSlim-MnasNet \cite{yu2019autoslim}& 33 ms\\
\hline 
\textbf{Dynamic OFA} & \textbf{73 ms}\\
\hline  
\end{tabular}}  \\
\caption{Comparison of average runtime DNN architecture switching time on GPU} 
\label{tab1}
\end{table}

At runtime, different sub-network architectures can be used to meet dynamic software performance requirements and available hardware resources. The runtime architecture switching time is shown in Table \ref{tab1}; Dynamic-OFA supports real-time DNN architecture switching, its switching time is much faster than the static OFA model \cite{cai2019once} since we only do search in a small subset of sub-networks. Furthermore, loading the pre-calculated batch-norm takes only 2ms. Dynamic-OFA is only slightly slower than AutoSlim\cite{yu2019autoslim} and MutualNet\cite{yang2020mutualnet} due to larger memory footprint, but it has a much better accuracy-latency/FLOPs trade-offs (Figs \ref{fig4} and \ref{fig5}). Such a small difference is not a considerable issue in real applications, since new operating environments (\ie software performance requirements, available hardware resources) normally last much longer (\eg minutes to hours).

\subsection{Dynamic Performance Requirements}
The sub-network architectures of Dynamic-OFA can be switched to meet the dynamic latency constraints at runtime. When only Dynamic-OFA is running on the device, all hardware resources are available to it. A look-up table that contains all accuracy-latency trade-offs can be obtained at design-time. RTMs can choose different operating points at runtime. As shown in Fig \ref{fig6}, the Dynamic-OFA model is deployed on the GPU and runs at the level 6 architecture while the latency constraint is 65ms at the beginning. At around 2500 ms, the latency constraint gets increased to 40 ms. Therefore the sub-network architecture is switched to level 2 for speedup at the cost of trading 4.13\% accuracy. Then, accuracy is recovered when the latency constraint is later reduced to 55ms.

The RTM, uses the sliding window to calculate the average latency  every 10 images. Based on the latency of our model on the GPU, the reaction time of the RTM is approximately 1-2s (\ie about 50 image classifications). The time overhead of the RTM is around 15 ms for each architecture switching, including the latency of monitoring 50 images and calculating the average time across the sliding windows.

\begin{figure}[t]
\begin{center}
   \includegraphics[width=0.9\linewidth]{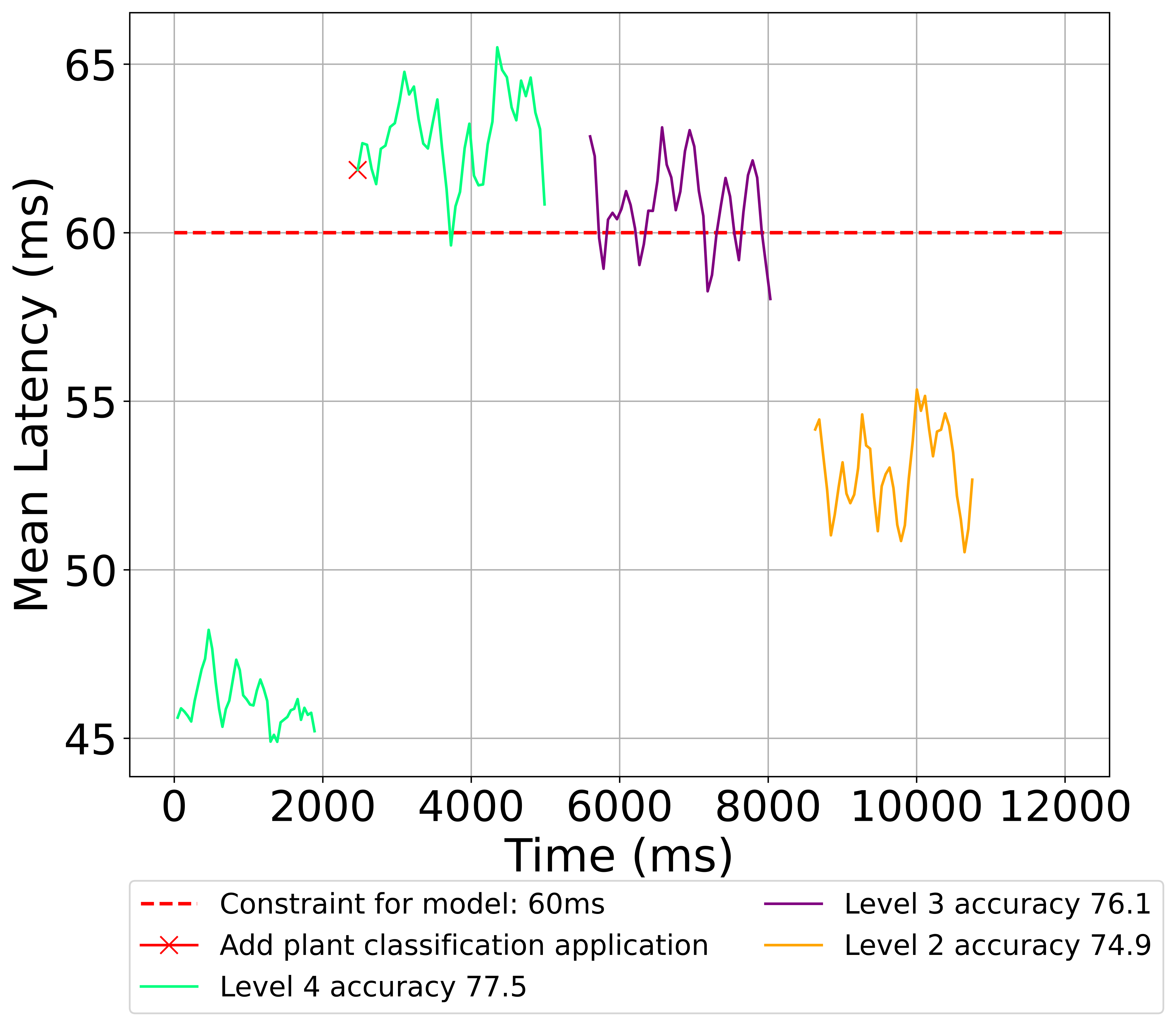}
\end{center}
   \caption{At $t=0$, a single DNN is executing. After ~2.5 s, a second application begins executing, reducing available GPU resource and impacting on the DNN inference latency. The runtime adapts, reducing the DNN accuracy until the performance constraint is met.}
\label{fig7}
\end{figure}

\subsection{Managing Concurrent Workloads}
The sub-network architectures of Dynamic-OFA can be switched to meet software performance constraints while fewer computing resources are available. Fig \ref{fig7} shows results where GPU computing resources are shared between Dynamic-OFA and a training task of a static DNN. The static DNN is an Nvidia open-source plant classification based on ResNet-18 \cite{PlantCLEF}. The training tasks starts to run at 2500 ms (donated by `X'), and Dynamic-OFA becomes slower since fewer GPU cores are available to it. The sub-network architecture is gradually switched from level 4 to level 2  to meet the latency constraint by trading 2.6\% accuracy.

\begin{figure}[t]
\begin{center}
   \includegraphics[width=0.9\linewidth]{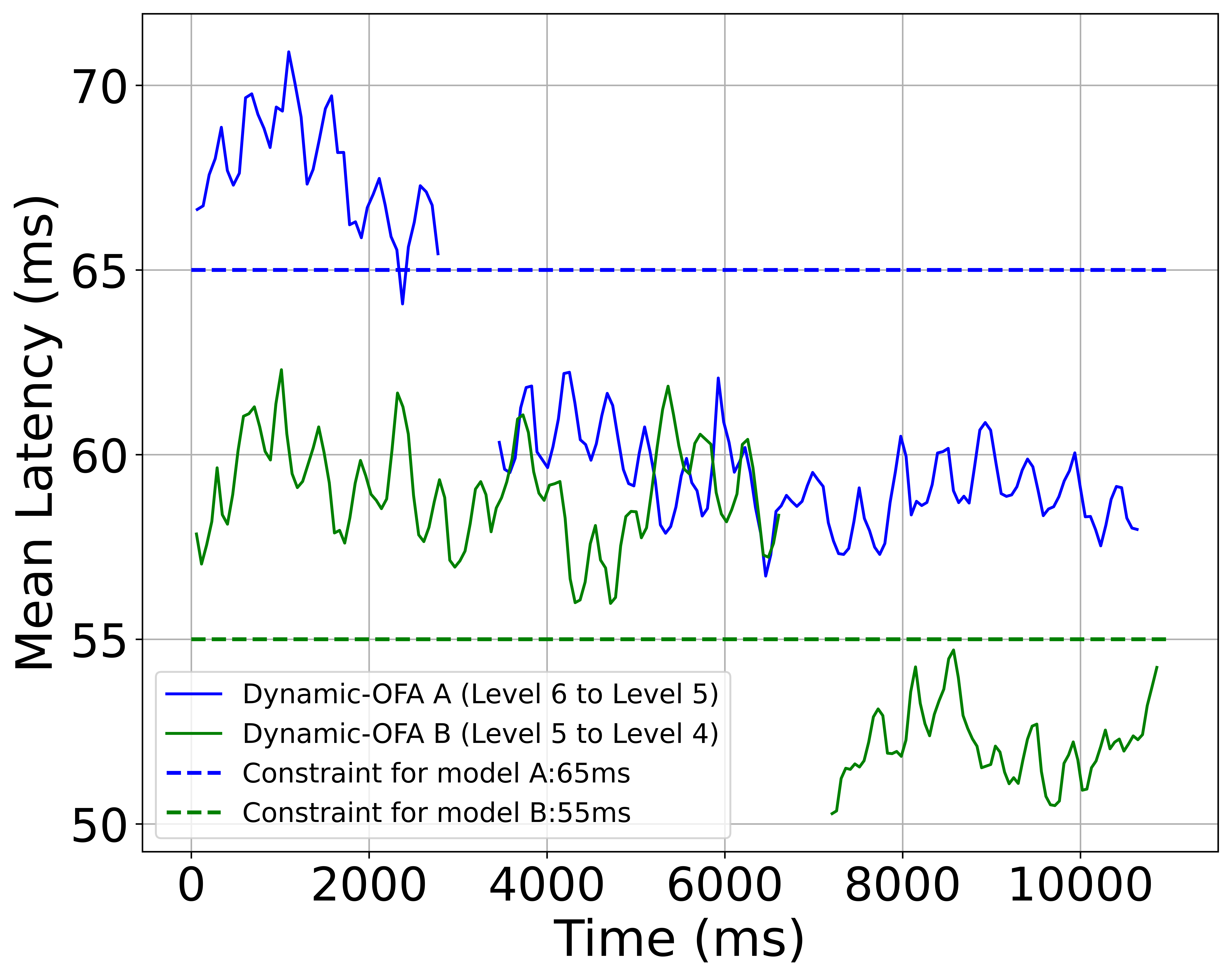}
\end{center}
   \caption{Runtime management of two concurrently executing Dynamic-OFA models, each with its own latency constraint (dashed lines).}
\label{fig8}
\end{figure}

Two Dynamic-OFA models can also coexist of when they share the same GPU. Fig \ref{fig8} shows two Dynamic-OFA models deployed on the same GPU, and their latency constraint become violated. The sub-network architecture of two Dynamic-OFA models are switched collectively so that the latency constraint of both models can be met, while keeping accuracy as high as possible. The constraints for model A and model B are 65 ms and 55 ms, respectively. Model A runs at the highest level at the beginning, and model B runs at level 5. To match the latency constraints, model A switches to level 5, but model B is still slower than the constraint, so model B switches to level 4: leaving both models meeting their constraints. 


\section{Conclusions}
This paper has proposed Dynamic-OFA, a novel dynamic DNN approach. Dynamic-OFA brings together two concepts: the OFA model and dynamic DNNs, which provide solid improvements over the previous state-of-the-art. Dynamic-OFA does not require any additional dynamic DNN model retraining and has the architecture flexibility for all heterogeneous computing elements with a share backbone. We empirically evaluated our approach against the start-of-the-art, our results show that our approach can provide better accuracy-latency trade-offs, up to 3.5x (CPU), 2.4x (GPU) faster for similar ImageNet Top-1 accuracy, or 3.8\% (CPU), 5.1\% (GPU) higher accuracy at similar latency.

Dynamic-OFA is a general approach for building dynamic DNNs, and the backbone network could be any super-networks trained by the OFA training pipeline. Our future work will investigate other applications such as network for IoT devices \cite{lin2020mcunet}, transformers for natural language processing (NLP) tasks \cite{hanruiwang2020hat}, generative adversarial networks (GANs) \cite{li2020gan}, 3D DNNs \cite{tang2020searching}, etc.

\section{Acknowledgements}
This work was supported in part by the Engineering and Physical Sciences Research Council (EPSRC) under Grant EP/S030069/1. Experimental data can be found at: https://doi.org/10.5258/SOTON/D1804. Code is available open-source on https://github.com/UoS-EEC.


\end{document}